\pdfoutput=1

\documentclass[11pt]{article}

\usepackage{etoolbox}
\apptocmd{\sloppy}{\hbadness 10000\relax}{}{}

\usepackage[table]{xcolor}
\usepackage{acl}
\usepackage{framed}
\usepackage{cancel}
\usepackage{graphicx}
\usepackage{subcaption}
\usepackage{booktabs}
\usepackage{tabularx}
\usepackage{multirow} 
\usepackage{longtable}
\usepackage[table]{xcolor}
\usepackage{adjustbox}

\usepackage{mdframed}

\usepackage{times}
\usepackage{latexsym}
\usepackage{booktabs}
\usepackage{amsmath}
\usepackage{svg}

\usepackage{graphicx}
\usepackage{subcaption}
\usepackage{booktabs} 
\usepackage{caption} 
\usepackage{array} 
\usepackage[T1]{fontenc}

\usepackage[utf8]{inputenc}

\usepackage{microtype}

\usepackage[disable]{todonotes}
\usepackage{comment}

\usepackage{xspace}

\title{Hypothesis-only Biases in Large Language Model-Elicited \\Natural Language Inference}

\author{Grace Proebsting \\
  \texttt{gproebstin@haverford.edu} \\\And
  Adam Poliak \\
  \texttt{apoliak@brynmawr.edu} \\}

\begin{document}
\maketitle
\begin{abstract}
We test whether replacing crowdsource workers with LLMs to write Natural Language Inference (NLI) hypotheses similarly results in annotation artifacts.
We recreate a portion of the Stanford NLI corpus using GPT-4, Llama-2 and Mistral 7b,
and train hypothesis-only classifiers to determine whether LLM-elicited hypotheses contain annotation artifacts. On our LLM-elicited NLI datasets, BERT-based hypothesis-only classifiers achieve between 86-96\% accuracy, indicating 
these datasets contain
hypothesis-only artifacts. We also find frequent ``give-aways'' in LLM-generated hypotheses, e.g. the phrase \textit{swimming in a pool} appears in more than 10,000 contradictions generated by GPT-4. Our analysis provides empirical evidence that well-attested biases in NLI can persist in LLM-generated data. 
\end{abstract}

\section{Introduction}
Creating NLP datasets with Large Language Models (LLMs) is an attractive alternative to relying on crowd-source  workers~\cite{ziems2023large}.
Compared to crowd-source workers, LLMs are inexpensive, fast, and always available. LLMs are an efficient tool for annotating data~\cite{zhao-etal-2022-lmturk,he2023annollm,bansal2023large,gilardi2023chatgpt} and generating training sets for NLP tasks~\cite{ye-etal-2022-zerogen,sahu-etal-2022-data,chung-etal-2023-increasing}.

We study whether LLM-elicited text suffers from \textit{annotation artifacts} similar to human crowd-sourced datasets~\cite{cai-etal-2017-pay,kaushik-lipton-2018-much}. We focus on Natural Language Inference (NLI), the task of determining whether a hypothesis sentence could be inferred from a premise~\cite{10.1007/11736790_9}, since popular crowd-sourced NLI datasets contain annotation artifacts. We create LLM-generated versions of the Stanford Natural Language Inference (SNLI) corpus~\cite{bowman-etal-2015-large} by providing LLMs with the same instructions given to SNLI crowd-source workers (\autoref{tab:example}).
We detect annotation artifacts in NLI using
hypothesis-only classifiers and identify give-away words associated with NLI labels. 

Our results demonstrate that without careful curation, LLM-elicited text contains annotation artifacts. We provide empirical evidence that well-attested biases in human-elicited NLI can persist in LLM-generated data, and that quality control is needed when generating datasets with LLMs.

\begin{table}[t!]
\footnotesize
\centering
\begin{tabularx}{\columnwidth}{lX} \toprule
\textbf{Premise} & Two women are hiking in the wilderness. \\\midrule
\multicolumn{2}{l}{\textbf{Entailment}} \\
\textbf{SNLI} & There are two women outdoors. \\
\textbf{Llama} & There are people outdoors. \\
\textbf{Mistral} & There are people in nature. \\
\textbf{GPT-4} & People are outdoors. \\\midrule
\multicolumn{2}{l}{\textbf{Contradiction}} \\
\textbf{SNLI} & There are two women in the living room. \\
\textbf{Llama} & A couple is having a picnic in a park. \\
\textbf{Mistral} & The women are shopping for clothes. \\
\textbf{GPT-4} & Two women are swimming in a pool. \\\bottomrule
\end{tabularx}
\caption{Entailed and contradicted hypotheses from humans (SNLI) and three LLMs (Llama-2 70b for Chat, Mistral 7b Instruct, and GPT-4) for the same premise.}
\label{tab:example}
\end{table}

\section{Related Work}
LLMs can efficiently generate, label, and clean datasets for a wide variety of applications~\cite{ziems2023large}. 
LLM-elicited text is especially attractive for sensitive domains, e.g. clinical NLP where datasets must not leak personal
identifying information~\cite{xu2023knowledgeinfused, frei2023annotated}.
LLMs have been used to generate instruction-tuning datasets \cite{honovich-etal-2023-unnatural,wang-etal-2023-self-instruct, peng2023instruction}, synthetic versions of benchmarks like SuperGLUE~\cite{wang2020superglue, gupta2023targen}, and counterfactuals for dataset augmentation~\cite{chen-etal-2023-disco,wu-etal-2021-polyjuice}. LLM-elicited text is pervasive even among crowd-source workers: ~\newcite{veselovsky2023artificial} claim that ``33–46\%" of the crowd-source workers hired for a summarization task likely used LLMs. 
Although many resources built with LLM-elicitation include quality assurance, such as ``human-in-the-loop'' curation ~\cite{kamalloo2023hagrid,wiegreffe-etal-2022-reframing, liu-etal-2022-wanli} or statistical filtering~\cite{wu-etal-2021-polyjuice,ye-etal-2022-zerogen,wiegreffe-etal-2022-reframing,wang-etal-2023-self-instruct,gupta2023targen,chen-etal-2023-disco,yehudai2024genie}, some LLM-generated datasets involve no post-filtering step ~\cite{peng2023instruction,xu2023contrastive,xu2023knowledgeinfused}.

\begin{table}[t!]
\center
\small
  \begin{tabular}{l r} 
    \toprule
\multicolumn{2}{l}{\textbf{Data set sizes:}}\\
Training pairs &  133,629\\
Evaluation pairs &  6,525\\
\midrule
\multicolumn{2}{l}{\textbf{Hypothesis mean token count:}}\\
SNLI train & 8.1 \\
Llama train & 9.4 \\
Mistral train & 9.1 \\
GPT-4 train & 9.2 \\
PaLM 2 train & 7.7 \\
\midrule
\multicolumn{2}{l}{\textbf{Mean Jaccard similarity with SNLI:}}\\
Llama train & 0.19 \\
Mistral train & 0.22 \\
GPT-4 train & 0.20 \\
PaLM 2 train & 0.25 \\
    \bottomrule
  \end{tabular}
\caption{Summary statistics for each dataset.}
\label{tab:collection-stats}
\end{table}

\section{Eliciting Hypotheses from LLMs} 
Prompting humans to generate text for large-scale NLP datasets can result in annotation artifacts~\cite{schwartz-etal-2017-effect,tsuchiya-2018-performance,gururangan-etal-2018-annotation,poliak-etal-2018-hypothesis,feng-etal-2019-misleading}. 
We ask whether LLM-elicited NLI hypotheses contain artifacts, and if so, what are they?

We create modified versions of SNLI by prompting LLMs with the same instructions that \newcite{bowman-etal-2015-large} provided crowd-source workers.
We verify the quality of the generated hypotheses and determine their similarity with hypotheses in SNLI.

\paragraph{LLMs under consideration}
We select a diverse set of LLMs for dataset generation: \textbf{GPT-4} \cite{openai2023gpt4}, \textbf{Llama-2 70b for Chat} \cite{touvron2023llama}, \textbf{Mistral 7b Instruct} \cite{jiang2023mistral}, and \textbf{PaLM 2 for Chat} \cite{anil2023palm}.\footnote{
 We use gpt-4-0613, llama-2-70b-chat,  mistral-7b-instruct-v0.1, chat-bison.} 
 These models vary in parameter count, parent company, and training technique.
We initially hoped to include models with open training sets to test for data contamination, e.g. AI2's OLMo-7B-Instruct~\cite{groeneveld2024olmo},
DataBrick's dolly-v2-12b~\cite{DatabricksBlog2023DollyV2} or EleutherAI’s gpt-j-6b~\cite{wang2021gpt}, but these open-data models did not create accurate entailed hypotheses in initial experiments. Given computational constraints, we were unable to use LLMs like BLOOM~\cite{workshop2022bloom} or Falcon-180b~\cite{almazrouei2023falcon}.

\paragraph{Dataset generation}
To fairly compare human- and LLM-elicited hypotheses, we prompted LLMs with the same instructions provided to crowd-source workers for SNLI.\footnote{We include the full prompt in the Appendix (\autoref{instructions-1}).} 
We set the temperature and top-p respectively to $0.75$ and $0.9$ for all LLMs to balance lexical diversity with reproducibility. Additionally, we use the default top-k parameter for each LLM. Due to budget constraints, for each LLM, we create hypotheses for a third of the premises in the SNLI train set and all premises in the SNLI dev set. \autoref{tab:collection-stats} reports statistics for each dataset.

\begin{table}[t!]
    \footnotesize
    \centering
    \begin{tabular}{c|c|c|c|c}
    \toprule
         &  Overall&  Entail&  Neutral& Contra\\ \midrule
         SNLI&  92.7&  87.0&  95.0& 96.0\\
         Llama &  89.7&  73.0&  98.0& 98.0\\
         Mistral &  83.7&  70.0&  91.0& 90.0\\
         GPT-4 &  94.3&  84.0&  99.0& 100.0\\
         PaLM 2&  77.0&  62.0&  90.0& 79.0\\ \bottomrule
    \end{tabular}
    \caption{Percentage of examples where we agreed with the label of 300 NLI example pairs from each dataset.} 
    \label{tab:validation}
\end{table}

\begin{figure}
    \centering
    \includegraphics[width=1\linewidth]{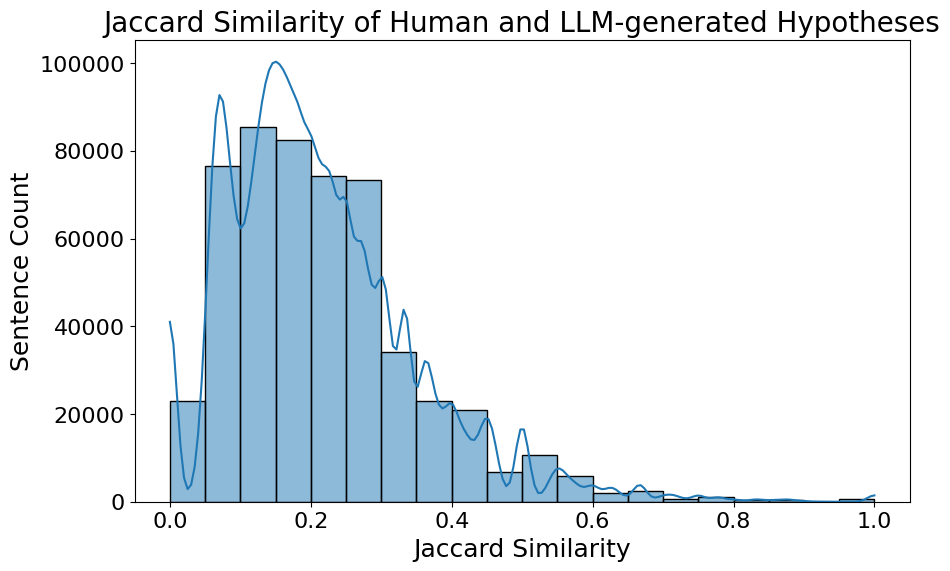}
    \caption{Frequency (y-axis) of lexical overlap (x-axis) between LLM and corresponding SNLI hypotheses. }
    \label{fig:jaccard}
\end{figure}

\paragraph{Dataset validation}
To verify that the LLMs correctly generated hypotheses for each label, we sampled 100 premises and manually checked the labels for the corresponding 300 NLI sentence pairs. \autoref{tab:validation} reports our agreement with the NLI labels for each LLM. We remove PaLM2 as we agreed with less than 80\% of sampled examples.

To ensure the LLM-generated hypotheses are not simply copied from SNLI, we compute the Jaccard similarity of the words within pairs of LLM-generated and SNLI hypotheses corresponding to the same premises and labels.
\autoref{fig:jaccard} shows that LLM and human-generated hypotheses have low lexical overlap, demonstrating that these LLMs do not copy hypotheses from SNLI verbatim. 

\section{Hypothesis-Only Classification}

\label{LLMs}
\begin{table}[t!]
    \centering
    \renewcommand{\arraystretch}{1.2}
    \begin{subtable}[t]{0.48\textwidth}
        \centering

        {\footnotesize
        \begin{tabular}{|c|c|c|c|c|}
            \hline
            \textbf{Train Set} & \multicolumn{4}{c|}{\textbf{Evaluation Set}} \\
            \cline{2-5}
            & SNLI & Llama & Mistral & GPT-4 \\
            \hline
            SNLI & 0.64 & 0.67 & 0.64 & 0.82 \\
            \hline
            Llama & 0.49 & 0.85 & 0.70 & 0.77 \\
            \hline
            Mistral & 0.45 & 0.61 & 0.80 & 0.64 \\
            \hline
            GPT-4 & 0.51 & 0.69 & 0.69 & 0.92 \\
            \hline
        \end{tabular}
                            \caption{Naive Bayes}
        \label{fig:naive_bayes}
        }
    \end{subtable}
    \hfill
    \begin{subtable}[t]{0.48\textwidth}
        \centering
        
        {\footnotesize
        \begin{tabular}{|c|c|c|c|c|}
            \hline
            \textbf{Train Set} & \multicolumn{4}{c|}{\textbf{Evaluation Set}} \\
            \cline{2-5}
            & SNLI & Llama & Mistral & GPT-4 \\
            \hline
            SNLI & 0.72 & 0.70 & 0.65 & 0.88 \\
            \hline
            Llama & 0.58 & 0.91 & 0.70 & 0.81 \\
            \hline
            Mistral & 0.51 & 0.77 & 0.86 & 0.72 \\
            \hline
            GPT-4 & 0.65 & 0.73 & 0.68 & 0.96 \\
            \hline
        \end{tabular}
                    \caption{BERT-based}  \label{fig:bert}
        }
    \end{subtable}
    \caption{Accuracy of each hypothesis-only classifier on each LLM and human-generated evaluation set. Each row represents the hypothesis-only dataset used for training, and each column represents the evaluation dataset.}
    \label{fig:hyp-only-llms}
\end{table}

We determine whether LLM-elicited NLI datasets contain annotation artifacts that allow a hypothesis-only model to significantly outperform a majority-class baseline.
We train two types of models: Naive Bayes (NB) using case-sensitive unigram features~\cite{scikit-learn}
and fine-tuned BERT-based classifier~\cite{DBLP:journals/corr/abs-1810-04805} with 3-class sequence classification heads using default Hugging Face hyper-parameters~\cite{wolf-etal-2020-transformers}.\footnote{We did not tune hyper-parameters
as our focus is identifying artifacts. We train the neural models for 1 epoch using AdamW \cite{loshchilov2019decoupled}, a learning rate of 2e-5, a weight decay of 0.01, and a batch size of 16.}

\paragraph{Accuracy of hypothesis-only classifiers}
We train hypothesis-only models on each of our train sets (3 LLM-generated and the filtered SNLI) and evaluate them on all evaluation sets.
\autoref{fig:hyp-only-llms} reports the accuracy of the hypothesis-only models. The hypothesis-only classifiers' high accuracy \textit{confirms the existence of hypothesis-only annotation artifacts in both human and LLM-generated NLI}. 

Surprisingly, the SNLI-trained models perform much better on the GPT-4 generated evaluation set (0.82 for NB and 0.88 for BERT) than on the SNLI evaluation set (0.64 for NB and 0.72 for BERT), indicating that GPT-4 might contain similar annotation artifacts as SNLI. We also notice that hypothesis-only models trained on LLM-generated data perform much better on other LLM-elicited datasets than on SNLI, as the accuracies in the first column are much lower than the other columns in both figures. This might indicate that the LLMs produce similar biases as each other.

\paragraph{Few unigrams needed for high NB accuracy.}

\begin{figure}[t!]
    \centering
     \includegraphics[width=\columnwidth]{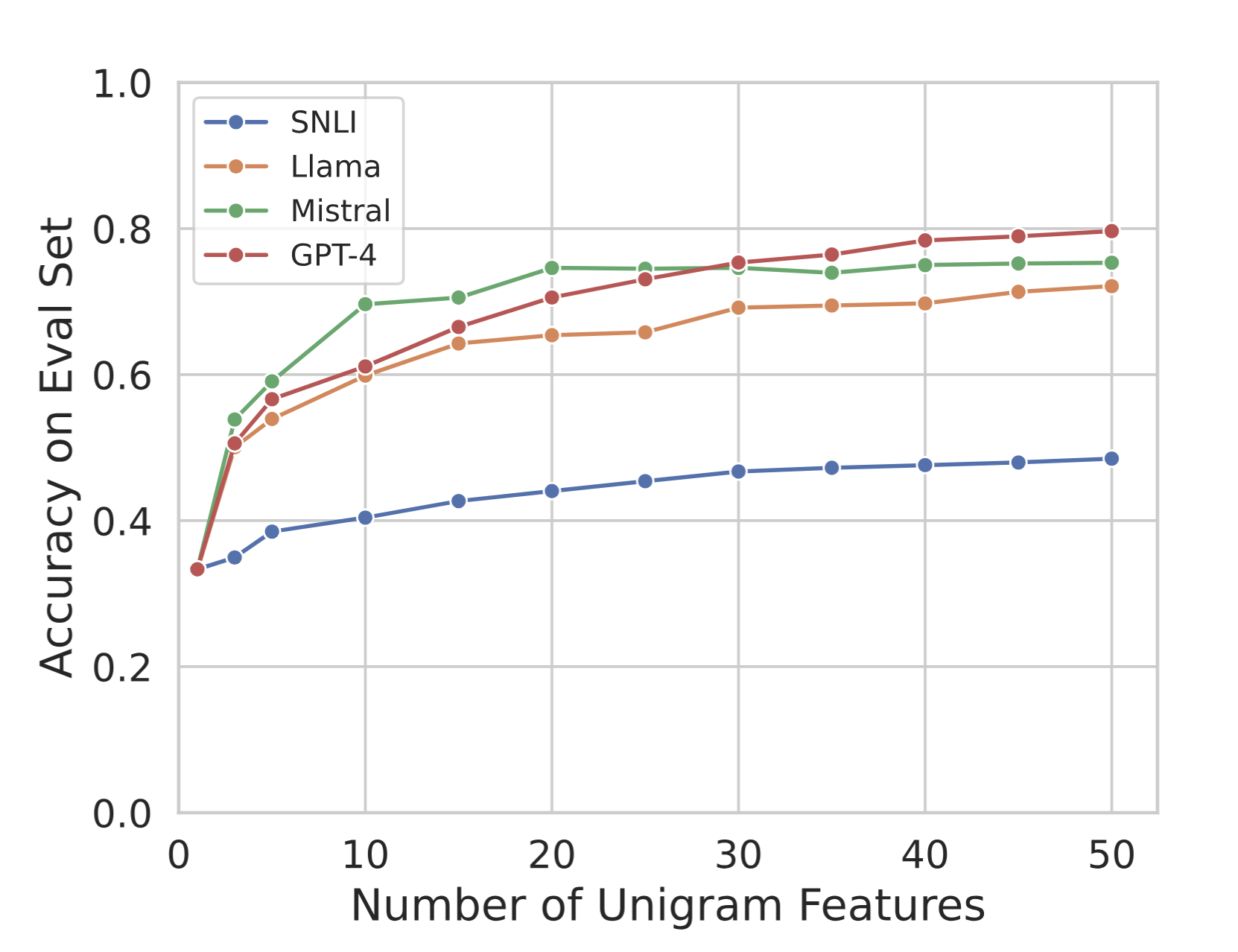}
    \caption{Accuracy of NB models using only the $n$ "most informative" unigram features for each train set evaluated on its corresponding evaluation set.}
    \label{fig:variable-features}
\end{figure}
The high accuracy of the NB models with unigram features indicates that the hypotheses contain \textit{give-away words}---single words that are highly indicative of a label. 

To determine how many give-away words are necessary to accurately classify LLM-elicited NLI, we train NB models based on the \textbf{n} most informative give-away words as features. We perform a chi-squared test on the words in each train set to determine informative give-away words. \autoref{fig:variable-features} reports the accuracy of NB hypothesis-only models using just 1 to 50 features. Compared to SNLI, the LLM-elicited datasets are far easier to classify using a sparse selection of unigram features. This result indicates that LLM-generated hypotheses are straightforward to classify not only due to the simplicity of the necessary features (unigrams) but also because only a negligibly small number of these simple features are required.

\begin{table*}[h!]
\small
\begin{adjustbox}{width=\textwidth}

\centering
\subfloat[][\textnormal{entailment}]{
\begin{tabular}[h]{l|c c c}

\toprule
& \textbf{Word} & \textbf{$p(l|w)$} & \textbf{Freq} \\
\midrule

& Humans & 0.95 & 128 \\
& least & 0.92 & 78 \\
SNLI & activity & 0.83 & 47 \\
& multiple & 0.81 & 37 \\
& interacting & 0.85 & 34 \\
\midrule 
& person & 0.81 & 22264 \\
& People & 0.86 & 7059 \\
Llama & standing & 0.84 & 4359 \\
& outdoors & 0.93 & 2390 \\
& engaging & 0.94 & 1689 \\
\midrule

& There & 0.99 & 16707 \\
& outdoors & 0.87 & 1055 \\
Mistral & three & 0.83 & 720 \\
& four & 0.88 & 335 \\
& urban & 0.83 & 318 \\
\midrule
& person & 0.85 & 11764 \\
& outdoors & 0.97 & 8182 \\
GPT-4 & individual & 0.96 & 4569 \\
& individuals & 0.89 & 3878 \\
& There & 0.86 & 3794 \\
\bottomrule
\end{tabular}
}
\subfloat[][\textnormal{neutral}]{
\begin{tabular}[h]{c c c}

\toprule
\textbf{Word} & \textbf{$p(l|w)$} & \textbf{Freq} \\
\midrule

tall & 0.85 & 418 \\
sad & 0.81 & 322 \\
first & 0.87 & 298 \\
owner & 0.83 & 284 \\
birthday & 0.83 & 227 \\
\midrule
Someone & 1 & 4092 \\
trying & 0.9 & 3023 \\
going & 0.95 & 1604 \\
break & 0.87 & 1339 \\
fun & 0.88 & 1165 \\
\midrule

be & 0.97 & 5154 \\
trying & 0.8 & 4875 \\
may & 0.98 & 3815 \\
having & 0.85 & 2039 \\
going & 0.83 & 1877 \\
\midrule

to & 0.85 & 7087 \\
for & 0.89 & 5791 \\
his & 0.82 & 5042 \\
friends & 0.94 & 3439 \\
enjoying & 0.85 & 2073 \\
\bottomrule
\end{tabular}
}
\subfloat[][\textnormal{contradiction}]{
\begin{tabular}[h]{c c c}

\toprule
\textbf{Word} & \textbf{$p(l|w)$} & \textbf{Freq} \\
\midrule

sleeping & 0.84 & 1747 \\
Nobody & 0.93 & 592 \\
asleep & 0.83 & 523 \\
couch & 0.81 & 477 \\
naked & 0.88 & 248 \\
\midrule
celebrity & 0.92 & 2359 \\
actually & 0.94 & 2075 \\
cat & 0.9 & 1973 \\
Everyone & 0.93 & 1913 \\
adult & 0.89 & 1782 \\
\midrule

The & 0.81 & 38491 \\
sitting & 0.83 & 14564 \\
bench & 0.87 & 8545 \\
not & 0.94 & 8068 \\
subject & 0.87 & 3672 \\
\midrule
swimming & 0.92 & 16281 \\
pool & 0.91 & 14638 \\
reading & 0.8 & 3492 \\
book & 0.81 & 3048 \\
sleeping & 0.91 & 2326 \\
\bottomrule
\end{tabular}
}
\end{adjustbox}
\caption{The most highly correlated words for each train set for given labels (the columns (c), (d), and (e)), thresholded to those with $p(l|w) >= 0.8$ and ranked according to frequency.}\label{top5-giveaway}
\end{table*}

\section{Qualitative Analysis}
We identify give-away words for each train set by calculating the conditional probability of each label $l$ given the presence of a word $w$ in a hypothesis. We consider all give-away words with a conditional probability of $\geq$ 0.8. \autoref{top5-giveaway} reports the top $5$ give-away words for each label in all train sets.\footnote{We include the top $10$ in the Appendix (\autoref{snli}).}

LLM-elicited give-away words seem to employ similar strategies or heuristics as human-elicited NLI.
Entailed examples in SNLI often contain generic words like \textit{humans}, \textit{activity}, and \textit{interacting}. We find a similar pattern in LLM-generated entailed hypotheses, e.g. \textit{person} and \textit{activity} in GPT-4 and Llama. Human-generated neutral hypotheses often contain modifiers (\textit{tall, sad, professional}) and superlatives (\textit{first, favorite, winning}). LLMs similarly add embellishing details about emotions or intentions (\textit{enjoying, fun, practicing, trying}) or the relationships between agents (\textit{friends, couple, team}) that are not explicit in the premise.

Both human- and LLM-elicited contradicting hypotheses contain negation words, e.g. \textit{nobody}, \textit{no}, \textit{not}.  
SNLI premises were ``sourced from Flickr naturally deal with activities''\cite{poliak-etal-2018-hypothesis},
so it is unsurprising that contradictions mention actions that cannot occur simultaneously with the action in the premise, e.g. \textit{sleeping} for SNLI, \textit{swimming} for GPT-4, and \textit{sitting} for Mistral. Further, LLMs frequently repeat contradictory \textit{phrases}: for instance, \textit{``swimming in a pool''} and \textit{``sitting on a bench''} each occur more than 10,000 times in the GPT-4 and Mistral-generated train sets.

Many give-away words for all three LLMs appear directly in the prompt. For example, \textit{There} is an entailment give-away for GPT-4, \textit{Someone} and \textit{catch} are neutral give-aways for Llama, and \textit{The}, \textit{sitting}, and \textit{couch} are contradiction give-aways for Mistral. This might be explained by the well-documented phenomena of LLMs often copying features in context~\cite{elhage2021mathematical, olsson2022context, bansal-etal-2023-rethinking,zhang2023benchmarking}. 

\section{Conclusion}
We studied whether Natural Language Inference datasets created by eliciting hypotheses from LLMs contain biases. We used $3$ LLMs to recreate a portion of SNLI and determined that LLM-elicited datasets contain annotation artifacts. Our analyses indicate that LLMs rely on similar strategies and heuristics as crowd-source workers when creating entailed, neutral, and contradicted hypotheses in response to a premise. 
Our findings provide a cautionary tale for relying on unfiltered, out-of-the-box LLM-generated data for classification datasets.

\bibliography{anthology,custom}

\begin{thebibliography}{49}
\expandafter\ifx\csname natexlab\endcsname\relax\def\natexlab#1{#1}\fi

\bibitem[{Almazrouei et~al.(2023)Almazrouei, Alobeidli, Alshamsi, Cappelli,
  Cojocaru, Debbah, Goffinet, Hesslow, Launay, Malartic
  et~al.}]{almazrouei2023falcon}
Ebtesam Almazrouei, Hamza Alobeidli, Abdulaziz Alshamsi, Alessandro Cappelli,
  Ruxandra Cojocaru, M{\'e}rouane Debbah, {\'E}tienne Goffinet, Daniel Hesslow,
  Julien Launay, Quentin Malartic, et~al. 2023.
\newblock The falcon series of open language models.
\newblock \emph{arXiv preprint arXiv:2311.16867}.

\bibitem[{Anil et~al.(2023)Anil, Dai, Firat, Johnson, Lepikhin, Passos,
  Shakeri, Taropa, Bailey, Chen, Chu, Clark, Shafey, Huang, Meier-Hellstern,
  Mishra, Moreira, Omernick, Robinson, Ruder, Tay, Xiao, Xu, Zhang, Abrego,
  Ahn, Austin, Barham, Botha, Bradbury, Brahma, Brooks, Catasta, Cheng, Cherry,
  Choquette-Choo, Chowdhery, Crepy, Dave, Dehghani, Dev, Devlin, Díaz, Du,
  Dyer, Feinberg, Feng, Fienber, Freitag, Garcia, Gehrmann, Gonzalez, Gur-Ari,
  Hand, Hashemi, Hou, Howland, Hu, Hui, Hurwitz, Isard, Ittycheriah, Jagielski,
  Jia, Kenealy, Krikun, Kudugunta, Lan, Lee, Lee, Li, Li, Li, Li, Li, Lim, Lin,
  Liu, Liu, Maggioni, Mahendru, Maynez, Misra, Moussalem, Nado, Nham, Ni,
  Nystrom, Parrish, Pellat, Polacek, Polozov, Pope, Qiao, Reif, Richter, Riley,
  Ros, Roy, Saeta, Samuel, Shelby, Slone, Smilkov, So, Sohn, Tokumine, Valter,
  Vasudevan, Vodrahalli, Wang, Wang, Wang, Wang, Wieting, Wu, Xu, Xu, Xue, Yin,
  Yu, Zhang, Zheng, Zheng, Zhou, Zhou, Petrov, and Wu}]{anil2023palm}
Rohan Anil, Andrew~M. Dai, Orhan Firat, Melvin Johnson, Dmitry Lepikhin,
  Alexandre Passos, Siamak Shakeri, Emanuel Taropa, Paige Bailey, Zhifeng Chen,
  Eric Chu, Jonathan~H. Clark, Laurent~El Shafey, Yanping Huang, Kathy
  Meier-Hellstern, Gaurav Mishra, Erica Moreira, Mark Omernick, Kevin Robinson,
  Sebastian Ruder, Yi~Tay, Kefan Xiao, Yuanzhong Xu, Yujing Zhang,
  Gustavo~Hernandez Abrego, Junwhan Ahn, Jacob Austin, Paul Barham, Jan Botha,
  James Bradbury, Siddhartha Brahma, Kevin Brooks, Michele Catasta, Yong Cheng,
  Colin Cherry, Christopher~A. Choquette-Choo, Aakanksha Chowdhery, Clément
  Crepy, Shachi Dave, Mostafa Dehghani, Sunipa Dev, Jacob Devlin, Mark Díaz,
  Nan Du, Ethan Dyer, Vlad Feinberg, Fangxiaoyu Feng, Vlad Fienber, Markus
  Freitag, Xavier Garcia, Sebastian Gehrmann, Lucas Gonzalez, Guy Gur-Ari,
  Steven Hand, Hadi Hashemi, Le~Hou, Joshua Howland, Andrea Hu, Jeffrey Hui,
  Jeremy Hurwitz, Michael Isard, Abe Ittycheriah, Matthew Jagielski, Wenhao
  Jia, Kathleen Kenealy, Maxim Krikun, Sneha Kudugunta, Chang Lan, Katherine
  Lee, Benjamin Lee, Eric Li, Music Li, Wei Li, YaGuang Li, Jian Li, Hyeontaek
  Lim, Hanzhao Lin, Zhongtao Liu, Frederick Liu, Marcello Maggioni, Aroma
  Mahendru, Joshua Maynez, Vedant Misra, Maysam Moussalem, Zachary Nado, John
  Nham, Eric Ni, Andrew Nystrom, Alicia Parrish, Marie Pellat, Martin Polacek,
  Alex Polozov, Reiner Pope, Siyuan Qiao, Emily Reif, Bryan Richter, Parker
  Riley, Alex~Castro Ros, Aurko Roy, Brennan Saeta, Rajkumar Samuel, Renee
  Shelby, Ambrose Slone, Daniel Smilkov, David~R. So, Daniel Sohn, Simon
  Tokumine, Dasha Valter, Vijay Vasudevan, Kiran Vodrahalli, Xuezhi Wang,
  Pidong Wang, Zirui Wang, Tao Wang, John Wieting, Yuhuai Wu, Kelvin Xu, Yunhan
  Xu, Linting Xue, Pengcheng Yin, Jiahui Yu, Qiao Zhang, Steven Zheng,
  Ce~Zheng, Weikang Zhou, Denny Zhou, Slav Petrov, and Yonghui Wu. 2023.
\newblock \href {http://arxiv.org/abs/2305.10403} {Palm 2 technical report}.

\bibitem[{Bansal et~al.(2023)Bansal, Gopalakrishnan, Dingliwal, Bodapati,
  Kirchhoff, and Roth}]{bansal-etal-2023-rethinking}
Hritik Bansal, Karthik Gopalakrishnan, Saket Dingliwal, Sravan Bodapati, Katrin
  Kirchhoff, and Dan Roth. 2023.
\newblock \href {https://aclanthology.org/2023.acl-long.660} {Rethinking the
  role of scale for in-context learning: An interpretability-based case study
  at 66 billion scale}.
\newblock In \emph{Proceedings of the 61st Annual Meeting of the Association
  for Computational Linguistics (Volume 1: Long Papers)}, pages 11833--11856,
  Toronto, Canada. Association for Computational Linguistics.

\bibitem[{Bansal and Sharma(2023)}]{bansal2023large}
Parikshit Bansal and Amit Sharma. 2023.
\newblock Large language models as annotators: Enhancing generalization of nlp
  models at minimal cost.
\newblock \emph{arXiv preprint arXiv:2306.15766}.

\bibitem[{Bowman et~al.(2015)Bowman, Angeli, Potts, and
  Manning}]{bowman-etal-2015-large}
Samuel~R. Bowman, Gabor Angeli, Christopher Potts, and Christopher~D. Manning.
  2015.
\newblock \href {https://doi.org/10.18653/v1/D15-1075} {A large annotated
  corpus for learning natural language inference}.
\newblock In \emph{Proceedings of the 2015 Conference on Empirical Methods in
  Natural Language Processing}, pages 632--642, Lisbon, Portugal. Association
  for Computational Linguistics.

\bibitem[{Cai et~al.(2017)Cai, Tu, and Gimpel}]{cai-etal-2017-pay}
Zheng Cai, Lifu Tu, and Kevin Gimpel. 2017.
\newblock \href {https://doi.org/10.18653/v1/P17-2097} {Pay attention to the
  ending:strong neural baselines for the {ROC} story cloze task}.
\newblock In \emph{Proceedings of the 55th Annual Meeting of the Association
  for Computational Linguistics (Volume 2: Short Papers)}, pages 616--622,
  Vancouver, Canada. Association for Computational Linguistics.

\bibitem[{Chen et~al.(2023)Chen, Gao, Bosselut, Sabharwal, and
  Richardson}]{chen-etal-2023-disco}
Zeming Chen, Qiyue Gao, Antoine Bosselut, Ashish Sabharwal, and Kyle
  Richardson. 2023.
\newblock \href {https://aclanthology.org/2023.acl-long.302} {{DISCO}:
  Distilling counterfactuals with large language models}.
\newblock In \emph{Proceedings of the 61st Annual Meeting of the Association
  for Computational Linguistics (Volume 1: Long Papers)}, pages 5514--5528,
  Toronto, Canada. Association for Computational Linguistics.

\bibitem[{Chung et~al.(2023)Chung, Kamar, and
  Amershi}]{chung-etal-2023-increasing}
John Chung, Ece Kamar, and Saleema Amershi. 2023.
\newblock \href {https://aclanthology.org/2023.acl-long.34} {Increasing
  diversity while maintaining accuracy: Text data generation with large
  language models and human interventions}.
\newblock In \emph{Proceedings of the 61st Annual Meeting of the Association
  for Computational Linguistics (Volume 1: Long Papers)}, pages 575--593,
  Toronto, Canada. Association for Computational Linguistics.

\bibitem[{Conover et~al.(2023)Conover, Hayes, Mathur, Xie, Wan, Shah, Ghodsi,
  Wendell, Zaharia, and Xin}]{DatabricksBlog2023DollyV2}
Mike Conover, Matt Hayes, Ankit Mathur, Jianwei Xie, Jun Wan, Sam Shah, Ali
  Ghodsi, Patrick Wendell, Matei Zaharia, and Reynold Xin. 2023.
\newblock \href
  {https://www.databricks.com/blog/2023/04/12/dolly-first-open-commercially-viable-instruction-tuned-llm}
  {Free dolly: Introducing the world's first truly open instruction-tuned llm}.

\bibitem[{Dagan et~al.(2005)Dagan, Glickman, and Magnini}]{10.1007/11736790_9}
Ido Dagan, Oren Glickman, and Bernardo Magnini. 2005.
\newblock \href {https://doi.org/10.1007/11736790_9} {The pascal recognising
  textual entailment challenge}.
\newblock In \emph{Proceedings of the First International Conference on Machine
  Learning Challenges: Evaluating Predictive Uncertainty Visual Object
  Classification, and Recognizing Textual Entailment}, MLCW'05, page 177–190,
  Berlin, Heidelberg. Springer-Verlag.

\bibitem[{Devlin et~al.(2019)Devlin, Chang, Lee, and
  Toutanova}]{DBLP:journals/corr/abs-1810-04805}
Jacob Devlin, Ming-Wei Chang, Kenton Lee, and Kristina Toutanova. 2019.
\newblock \href {https://doi.org/10.18653/v1/N19-1423} {{BERT}: Pre-training of
  deep bidirectional transformers for language understanding}.
\newblock In \emph{Proceedings of the 2019 Conference of the North {A}merican
  Chapter of the Association for Computational Linguistics: Human Language
  Technologies, Volume 1 (Long and Short Papers)}, pages 4171--4186,
  Minneapolis, Minnesota. Association for Computational Linguistics.

\bibitem[{Elhage et~al.(2021)Elhage, Nanda, Olsson, Henighan, Joseph, Mann,
  Askell, Bai, Chen, Conerly, DasSarma, Drain, Ganguli, Hatfield-Dodds,
  Hernandez, Jones, Kernion, Lovitt, Ndousse, Amodei, Brown, Clark, Kaplan,
  McCandlish, and Olah}]{elhage2021mathematical}
Nelson Elhage, Neel Nanda, Catherine Olsson, Tom Henighan, Nicholas Joseph, Ben
  Mann, Amanda Askell, Yuntao Bai, Anna Chen, Tom Conerly, Nova DasSarma, Dawn
  Drain, Deep Ganguli, Zac Hatfield-Dodds, Danny Hernandez, Andy Jones, Jackson
  Kernion, Liane Lovitt, Kamal Ndousse, Dario Amodei, Tom Brown, Jack Clark,
  Jared Kaplan, Sam McCandlish, and Chris Olah. 2021.
\newblock A mathematical framework for transformer circuits.
\newblock \emph{Transformer Circuits Thread}.
\newblock Https://transformer-circuits.pub/2021/framework/index.html.

\bibitem[{Feng et~al.(2019)Feng, Wallace, and
  Boyd-Graber}]{feng-etal-2019-misleading}
Shi Feng, Eric Wallace, and Jordan Boyd-Graber. 2019.
\newblock \href {https://doi.org/10.18653/v1/P19-1554} {Misleading failures of
  partial-input baselines}.
\newblock In \emph{Proceedings of the 57th Annual Meeting of the Association
  for Computational Linguistics}, pages 5533--5538, Florence, Italy.
  Association for Computational Linguistics.

\bibitem[{Frei and Kramer(2023)}]{frei2023annotated}
Johann Frei and Frank Kramer. 2023.
\newblock Annotated dataset creation through large language models for
  non-english medical nlp.
\newblock \emph{Journal of Biomedical Informatics}, page 104478.

\bibitem[{Gilardi et~al.(2023)Gilardi, Alizadeh, and
  Kubli}]{gilardi2023chatgpt}
Fabrizio Gilardi, Meysam Alizadeh, and Ma{\"e}l Kubli. 2023.
\newblock Chatgpt outperforms crowd-workers for text-annotation tasks.
\newblock \emph{arXiv preprint arXiv:2303.15056}.

\bibitem[{Groeneveld et~al.(2024)Groeneveld, Beltagy, Walsh, Bhagia, Kinney,
  Tafjord, Jha, Ivison, Magnusson, Wang, Arora, Atkinson, Authur, Chandu,
  Cohan, Dumas, Elazar, Gu, Hessel, Khot, Merrill, Morrison, Muennighoff, Naik,
  Nam, Peters, Pyatkin, Ravichander, Schwenk, Shah, Smith, Strubell, Subramani,
  Wortsman, Dasigi, Lambert, Richardson, Zettlemoyer, Dodge, Lo, Soldaini,
  Smith, and Hajishirzi}]{groeneveld2024olmo}
Dirk Groeneveld, Iz~Beltagy, Pete Walsh, Akshita Bhagia, Rodney Kinney, Oyvind
  Tafjord, Ananya~Harsh Jha, Hamish Ivison, Ian Magnusson, Yizhong Wang, Shane
  Arora, David Atkinson, Russell Authur, Khyathi~Raghavi Chandu, Arman Cohan,
  Jennifer Dumas, Yanai Elazar, Yuling Gu, Jack Hessel, Tushar Khot, William
  Merrill, Jacob Morrison, Niklas Muennighoff, Aakanksha Naik, Crystal Nam,
  Matthew~E. Peters, Valentina Pyatkin, Abhilasha Ravichander, Dustin Schwenk,
  Saurabh Shah, Will Smith, Emma Strubell, Nishant Subramani, Mitchell
  Wortsman, Pradeep Dasigi, Nathan Lambert, Kyle Richardson, Luke Zettlemoyer,
  Jesse Dodge, Kyle Lo, Luca Soldaini, Noah~A. Smith, and Hannaneh Hajishirzi.
  2024.
\newblock \href {http://arxiv.org/abs/2402.00838} {Olmo: Accelerating the
  science of language models}.

\bibitem[{Gupta et~al.(2023)Gupta, Scaria, Anantheswaran, Verma, Parmar,
  Sawant, Mishra, and Baral}]{gupta2023targen}
Himanshu Gupta, Kevin Scaria, Ujjwala Anantheswaran, Shreyas Verma, Mihir
  Parmar, Saurabh~Arjun Sawant, Swaroop Mishra, and Chitta Baral. 2023.
\newblock Targen: Targeted data generation with large language models.
\newblock \emph{arXiv preprint arXiv:2310.17876}.

\bibitem[{Gururangan et~al.(2018)Gururangan, Swayamdipta, Levy, Schwartz,
  Bowman, and Smith}]{gururangan-etal-2018-annotation}
Suchin Gururangan, Swabha Swayamdipta, Omer Levy, Roy Schwartz, Samuel Bowman,
  and Noah~A. Smith. 2018.
\newblock \href {https://doi.org/10.18653/v1/N18-2017} {Annotation artifacts in
  natural language inference data}.
\newblock In \emph{Proceedings of the 2018 Conference of the North {A}merican
  Chapter of the Association for Computational Linguistics: Human Language
  Technologies, Volume 2 (Short Papers)}, pages 107--112, New Orleans,
  Louisiana. Association for Computational Linguistics.

\bibitem[{He et~al.(2023)He, Lin, Gong, Jin, Zhang, Lin, Jiao, Yiu, Duan, Chen
  et~al.}]{he2023annollm}
Xingwei He, Zhenghao Lin, Yeyun Gong, Alex Jin, Hang Zhang, Chen Lin, Jian
  Jiao, Siu~Ming Yiu, Nan Duan, Weizhu Chen, et~al. 2023.
\newblock Annollm: Making large language models to be better crowdsourced
  annotators.
\newblock \emph{arXiv preprint arXiv:2303.16854}.

\bibitem[{Honovich et~al.(2023)Honovich, Scialom, Levy, and
  Schick}]{honovich-etal-2023-unnatural}
Or~Honovich, Thomas Scialom, Omer Levy, and Timo Schick. 2023.
\newblock \href {https://aclanthology.org/2023.acl-long.806} {Unnatural
  instructions: Tuning language models with (almost) no human labor}.
\newblock In \emph{Proceedings of the 61st Annual Meeting of the Association
  for Computational Linguistics (Volume 1: Long Papers)}, pages 14409--14428,
  Toronto, Canada. Association for Computational Linguistics.

\bibitem[{Jiang et~al.(2023)Jiang, Sablayrolles, Mensch, Bamford, Chaplot,
  de~las Casas, Bressand, Lengyel, Lample, Saulnier, Lavaud, Lachaux, Stock,
  Scao, Lavril, Wang, Lacroix, and Sayed}]{jiang2023mistral}
Albert~Q. Jiang, Alexandre Sablayrolles, Arthur Mensch, Chris Bamford,
  Devendra~Singh Chaplot, Diego de~las Casas, Florian Bressand, Gianna Lengyel,
  Guillaume Lample, Lucile Saulnier, Lélio~Renard Lavaud, Marie-Anne Lachaux,
  Pierre Stock, Teven~Le Scao, Thibaut Lavril, Thomas Wang, Timothée Lacroix,
  and William~El Sayed. 2023.
\newblock \href {http://arxiv.org/abs/2310.06825} {Mistral 7b}.

\bibitem[{Kamalloo et~al.(2023)Kamalloo, Jafari, Zhang, Thakur, and
  Lin}]{kamalloo2023hagrid}
Ehsan Kamalloo, Aref Jafari, Xinyu Zhang, Nandan Thakur, and Jimmy Lin. 2023.
\newblock \href {http://arxiv.org/abs/2307.16883} {Hagrid: A human-llm
  collaborative dataset for generative information-seeking with attribution}.

\bibitem[{Kaushik and Lipton(2018)}]{kaushik-lipton-2018-much}
Divyansh Kaushik and Zachary~C. Lipton. 2018.
\newblock \href {https://doi.org/10.18653/v1/D18-1546} {How much reading does
  reading comprehension require? a critical investigation of popular
  benchmarks}.
\newblock In \emph{Proceedings of the 2018 Conference on Empirical Methods in
  Natural Language Processing}, pages 5010--5015, Brussels, Belgium.
  Association for Computational Linguistics.

\bibitem[{Liu et~al.(2022)Liu, Swayamdipta, Smith, and
  Choi}]{liu-etal-2022-wanli}
Alisa Liu, Swabha Swayamdipta, Noah~A. Smith, and Yejin Choi. 2022.
\newblock \href {https://aclanthology.org/2022.findings-emnlp.508} {{WANLI}:
  Worker and {AI} collaboration for natural language inference dataset
  creation}.
\newblock In \emph{Findings of the Association for Computational Linguistics:
  EMNLP 2022}, pages 6826--6847, Abu Dhabi, United Arab Emirates. Association
  for Computational Linguistics.

\bibitem[{Loshchilov and Hutter(2018)}]{loshchilov2019decoupled}
Ilya Loshchilov and Frank Hutter. 2018.
\newblock Decoupled weight decay regularization.
\newblock In \emph{International Conference on Learning Representations}.

\bibitem[{Olsson et~al.(2022)Olsson, Elhage, Nanda, Joseph, DasSarma, Henighan,
  Mann, Askell, Bai, Chen, Conerly, Drain, Ganguli, Hatfield-Dodds, Hernandez,
  Johnston, Jones, Kernion, Lovitt, Ndousse, Amodei, Brown, Clark, Kaplan,
  McCandlish, and Olah}]{olsson2022context}
Catherine Olsson, Nelson Elhage, Neel Nanda, Nicholas Joseph, Nova DasSarma,
  Tom Henighan, Ben Mann, Amanda Askell, Yuntao Bai, Anna Chen, Tom Conerly,
  Dawn Drain, Deep Ganguli, Zac Hatfield-Dodds, Danny Hernandez, Scott
  Johnston, Andy Jones, Jackson Kernion, Liane Lovitt, Kamal Ndousse, Dario
  Amodei, Tom Brown, Jack Clark, Jared Kaplan, Sam McCandlish, and Chris Olah.
  2022.
\newblock In-context learning and induction heads.
\newblock \emph{Transformer Circuits Thread}.
\newblock
  Https://transformer-circuits.pub/2022/in-context-learning-and-induction-heads/index.html.

\bibitem[{OpenAI(2023)}]{openai2023gpt4}
OpenAI. 2023.
\newblock \href {http://arxiv.org/abs/2303.08774} {Gpt-4 technical report}.

\bibitem[{Pedregosa et~al.(2011)Pedregosa, Varoquaux, Gramfort, Michel,
  Thirion, Grisel, Blondel, Prettenhofer, Weiss, Dubourg, Vanderplas, Passos,
  Cournapeau, Brucher, Perrot, and Duchesnay}]{scikit-learn}
F.~Pedregosa, G.~Varoquaux, A.~Gramfort, V.~Michel, B.~Thirion, O.~Grisel,
  M.~Blondel, P.~Prettenhofer, R.~Weiss, V.~Dubourg, J.~Vanderplas, A.~Passos,
  D.~Cournapeau, M.~Brucher, M.~Perrot, and E.~Duchesnay. 2011.
\newblock Scikit-learn: Machine learning in {P}ython.
\newblock \emph{Journal of Machine Learning Research}, 12:2825--2830.

\bibitem[{Peng et~al.(2023)Peng, Li, He, Galley, and Gao}]{peng2023instruction}
Baolin Peng, Chunyuan Li, Pengcheng He, Michel Galley, and Jianfeng Gao. 2023.
\newblock \href {http://arxiv.org/abs/2304.03277} {Instruction tuning with
  gpt-4}.

\bibitem[{Poliak et~al.(2018)Poliak, Naradowsky, Haldar, Rudinger, and
  Van~Durme}]{poliak-etal-2018-hypothesis}
Adam Poliak, Jason Naradowsky, Aparajita Haldar, Rachel Rudinger, and Benjamin
  Van~Durme. 2018.
\newblock \href {https://doi.org/10.18653/v1/S18-2023} {Hypothesis only
  baselines in natural language inference}.
\newblock In \emph{Proceedings of the Seventh Joint Conference on Lexical and
  Computational Semantics}, pages 180--191, New Orleans, Louisiana. Association
  for Computational Linguistics.

\bibitem[{Sahu et~al.(2022)Sahu, Rodriguez, Laradji, Atighehchian, Vazquez, and
  Bahdanau}]{sahu-etal-2022-data}
Gaurav Sahu, Pau Rodriguez, Issam Laradji, Parmida Atighehchian, David Vazquez,
  and Dzmitry Bahdanau. 2022.
\newblock \href {https://doi.org/10.18653/v1/2022.nlp4convai-1.5} {Data
  augmentation for intent classification with off-the-shelf large language
  models}.
\newblock In \emph{Proceedings of the 4th Workshop on NLP for Conversational
  AI}, pages 47--57, Dublin, Ireland. Association for Computational
  Linguistics.

\bibitem[{Schwartz et~al.(2017)Schwartz, Sap, Konstas, Zilles, Choi, and
  Smith}]{schwartz-etal-2017-effect}
Roy Schwartz, Maarten Sap, Ioannis Konstas, Leila Zilles, Yejin Choi, and
  Noah~A. Smith. 2017.
\newblock \href {https://doi.org/10.18653/v1/K17-1004} {The effect of different
  writing tasks on linguistic style: A case study of the {ROC} story cloze
  task}.
\newblock In \emph{Proceedings of the 21st Conference on Computational Natural
  Language Learning ({C}o{NLL} 2017)}, pages 15--25, Vancouver, Canada.
  Association for Computational Linguistics.

\bibitem[{Touvron et~al.(2023)Touvron, Martin, Stone, Albert, Almahairi,
  Babaei, Bashlykov, Batra, Bhargava, Bhosale, Bikel, Blecher, Ferrer, Chen,
  Cucurull, Esiobu, Fernandes, Fu, Fu, Fuller, Gao, Goswami, Goyal, Hartshorn,
  Hosseini, Hou, Inan, Kardas, Kerkez, Khabsa, Kloumann, Korenev, Koura,
  Lachaux, Lavril, Lee, Liskovich, Lu, Mao, Martinet, Mihaylov, Mishra,
  Molybog, Nie, Poulton, Reizenstein, Rungta, Saladi, Schelten, Silva, Smith,
  Subramanian, Tan, Tang, Taylor, Williams, Kuan, Xu, Yan, Zarov, Zhang, Fan,
  Kambadur, Narang, Rodriguez, Stojnic, Edunov, and Scialom}]{touvron2023llama}
Hugo Touvron, Louis Martin, Kevin Stone, Peter Albert, Amjad Almahairi, Yasmine
  Babaei, Nikolay Bashlykov, Soumya Batra, Prajjwal Bhargava, Shruti Bhosale,
  Dan Bikel, Lukas Blecher, Cristian~Canton Ferrer, Moya Chen, Guillem
  Cucurull, David Esiobu, Jude Fernandes, Jeremy Fu, Wenyin Fu, Brian Fuller,
  Cynthia Gao, Vedanuj Goswami, Naman Goyal, Anthony Hartshorn, Saghar
  Hosseini, Rui Hou, Hakan Inan, Marcin Kardas, Viktor Kerkez, Madian Khabsa,
  Isabel Kloumann, Artem Korenev, Punit~Singh Koura, Marie-Anne Lachaux,
  Thibaut Lavril, Jenya Lee, Diana Liskovich, Yinghai Lu, Yuning Mao, Xavier
  Martinet, Todor Mihaylov, Pushkar Mishra, Igor Molybog, Yixin Nie, Andrew
  Poulton, Jeremy Reizenstein, Rashi Rungta, Kalyan Saladi, Alan Schelten, Ruan
  Silva, Eric~Michael Smith, Ranjan Subramanian, Xiaoqing~Ellen Tan, Binh Tang,
  Ross Taylor, Adina Williams, Jian~Xiang Kuan, Puxin Xu, Zheng Yan, Iliyan
  Zarov, Yuchen Zhang, Angela Fan, Melanie Kambadur, Sharan Narang, Aurelien
  Rodriguez, Robert Stojnic, Sergey Edunov, and Thomas Scialom. 2023.
\newblock \href {http://arxiv.org/abs/2307.09288} {Llama 2: Open foundation and
  fine-tuned chat models}.

\bibitem[{Tsuchiya(2018)}]{tsuchiya-2018-performance}
Masatoshi Tsuchiya. 2018.
\newblock \href {https://aclanthology.org/L18-1239} {Performance impact caused
  by hidden bias of training data for recognizing textual entailment}.
\newblock In \emph{Proceedings of the Eleventh International Conference on
  Language Resources and Evaluation ({LREC} 2018)}, Miyazaki, Japan. European
  Language Resources Association (ELRA).

\bibitem[{Veselovsky et~al.(2023)Veselovsky, Ribeiro, and
  West}]{veselovsky2023artificial}
Veniamin Veselovsky, Manoel~Horta Ribeiro, and Robert West. 2023.
\newblock Artificial artificial artificial intelligence: Crowd workers widely
  use large language models for text production tasks.
\newblock \emph{arXiv preprint arXiv:2306.07899}.

\bibitem[{Wang et~al.(2019)Wang, Pruksachatkun, Nangia, Singh, Michael, Hill,
  Levy, and Bowman}]{wang2020superglue}
Alex Wang, Yada Pruksachatkun, Nikita Nangia, Amanpreet Singh, Julian Michael,
  Felix Hill, Omer Levy, and Samuel Bowman. 2019.
\newblock Superglue: A stickier benchmark for general-purpose language
  understanding systems.
\newblock \emph{Advances in neural information processing systems}, 32.

\bibitem[{Wang and Komatsuzaki(2021)}]{wang2021gpt}
Ben Wang and Aran Komatsuzaki. 2021.
\newblock Gpt-j-6b: A 6 billion parameter autoregressive language model.

\bibitem[{Wang et~al.(2023)Wang, Kordi, Mishra, Liu, Smith, Khashabi, and
  Hajishirzi}]{wang-etal-2023-self-instruct}
Yizhong Wang, Yeganeh Kordi, Swaroop Mishra, Alisa Liu, Noah~A. Smith, Daniel
  Khashabi, and Hannaneh Hajishirzi. 2023.
\newblock \href {https://aclanthology.org/2023.acl-long.754} {Self-instruct:
  Aligning language models with self-generated instructions}.
\newblock In \emph{Proceedings of the 61st Annual Meeting of the Association
  for Computational Linguistics (Volume 1: Long Papers)}, pages 13484--13508,
  Toronto, Canada. Association for Computational Linguistics.

\bibitem[{Wiegreffe et~al.(2022)Wiegreffe, Hessel, Swayamdipta, Riedl, and
  Choi}]{wiegreffe-etal-2022-reframing}
Sarah Wiegreffe, Jack Hessel, Swabha Swayamdipta, Mark Riedl, and Yejin Choi.
  2022.
\newblock \href {https://doi.org/10.18653/v1/2022.naacl-main.47} {Reframing
  human-{AI} collaboration for generating free-text explanations}.
\newblock In \emph{Proceedings of the 2022 Conference of the North American
  Chapter of the Association for Computational Linguistics: Human Language
  Technologies}, pages 632--658, Seattle, United States. Association for
  Computational Linguistics.

\bibitem[{Wolf et~al.(2020)Wolf, Debut, Sanh, Chaumond, Delangue, Moi, Cistac,
  Rault, Louf, Funtowicz, Davison, Shleifer, von Platen, Ma, Jernite, Plu, Xu,
  Le~Scao, Gugger, Drame, Lhoest, and Rush}]{wolf-etal-2020-transformers}
Thomas Wolf, Lysandre Debut, Victor Sanh, Julien Chaumond, Clement Delangue,
  Anthony Moi, Pierric Cistac, Tim Rault, Remi Louf, Morgan Funtowicz, Joe
  Davison, Sam Shleifer, Patrick von Platen, Clara Ma, Yacine Jernite, Julien
  Plu, Canwen Xu, Teven Le~Scao, Sylvain Gugger, Mariama Drame, Quentin Lhoest,
  and Alexander Rush. 2020.
\newblock \href {https://doi.org/10.18653/v1/2020.emnlp-demos.6} {Transformers:
  State-of-the-art natural language processing}.
\newblock In \emph{Proceedings of the 2020 Conference on Empirical Methods in
  Natural Language Processing: System Demonstrations}, pages 38--45, Online.
  Association for Computational Linguistics.

\bibitem[{Workshop et~al.(2022)Workshop, Scao, Fan, Akiki, Pavlick, Ili{\'c},
  Hesslow, Castagn{\'e}, Luccioni, Yvon et~al.}]{workshop2022bloom}
BigScience Workshop, Teven~Le Scao, Angela Fan, Christopher Akiki, Ellie
  Pavlick, Suzana Ili{\'c}, Daniel Hesslow, Roman Castagn{\'e}, Alexandra~Sasha
  Luccioni, Fran{\c{c}}ois Yvon, et~al. 2022.
\newblock Bloom: A 176b-parameter open-access multilingual language model.
\newblock \emph{arXiv preprint arXiv:2211.05100}.

\bibitem[{Wu et~al.(2021)Wu, Ribeiro, Heer, and Weld}]{wu-etal-2021-polyjuice}
Tongshuang Wu, Marco~Tulio Ribeiro, Jeffrey Heer, and Daniel Weld. 2021.
\newblock \href {https://doi.org/10.18653/v1/2021.acl-long.523} {Polyjuice:
  Generating counterfactuals for explaining, evaluating, and improving models}.
\newblock In \emph{Proceedings of the 59th Annual Meeting of the Association
  for Computational Linguistics and the 11th International Joint Conference on
  Natural Language Processing (Volume 1: Long Papers)}, pages 6707--6723,
  Online. Association for Computational Linguistics.

\bibitem[{Xu et~al.(2023{\natexlab{a}})Xu, Rosset, Corro, Mahajan, McAuley,
  Neville, Awadallah, and Rao}]{xu2023contrastive}
Canwen Xu, Corby Rosset, Luciano~Del Corro, Shweti Mahajan, Julian McAuley,
  Jennifer Neville, Ahmed~Hassan Awadallah, and Nikhil Rao. 2023{\natexlab{a}}.
\newblock \href {http://arxiv.org/abs/2310.02263} {Contrastive post-training
  large language models on data curriculum}.

\bibitem[{Xu et~al.(2023{\natexlab{b}})Xu, Cui, Yu, Kan, Shi, Zhuang, Jin, Ho,
  and Yang}]{xu2023knowledgeinfused}
Ran Xu, Hejie Cui, Yue Yu, Xuan Kan, Wenqi Shi, Yuchen Zhuang, Wei Jin, Joyce
  Ho, and Carl Yang. 2023{\natexlab{b}}.
\newblock \href {http://arxiv.org/abs/2311.00287} {Knowledge-infused prompting:
  Assessing and advancing clinical text data generation with large language
  models}.

\bibitem[{Ye et~al.(2022)Ye, Gao, Li, Xu, Feng, Wu, Yu, and
  Kong}]{ye-etal-2022-zerogen}
Jiacheng Ye, Jiahui Gao, Qintong Li, Hang Xu, Jiangtao Feng, Zhiyong Wu, Tao
  Yu, and Lingpeng Kong. 2022.
\newblock \href {https://aclanthology.org/2022.emnlp-main.801} {{Z}ero{G}en:
  Efficient zero-shot learning via dataset generation}.
\newblock In \emph{Proceedings of the 2022 Conference on Empirical Methods in
  Natural Language Processing}, pages 11653--11669, Abu Dhabi, United Arab
  Emirates. Association for Computational Linguistics.

\bibitem[{Yehudai et~al.(2024)Yehudai, Carmeli, Mass, Arviv, Mills, Toledo,
  Shnarch, and Choshen}]{yehudai2024genie}
Asaf Yehudai, Boaz Carmeli, Yosi Mass, Ofir Arviv, Nathaniel Mills, Assaf
  Toledo, Eyal Shnarch, and Leshem Choshen. 2024.
\newblock Genie: Achieving human parity in content-grounded datasets
  generation.
\newblock \emph{arXiv preprint arXiv:2401.14367}.

\bibitem[{Zhang et~al.(2024)Zhang, Ladhak, Durmus, Liang, McKeown, and
  Hashimoto}]{zhang2023benchmarking}
Tianyi Zhang, Faisal Ladhak, Esin Durmus, Percy Liang, Kathleen McKeown, and
  Tatsunori~B. Hashimoto. 2024.
\newblock \href {https://doi.org/10.1162/tacl_a_00632} {{Benchmarking Large
  Language Models for News Summarization}}.
\newblock \emph{Transactions of the Association for Computational Linguistics},
  12:39--57.

\bibitem[{Zhao et~al.(2022)Zhao, Mi, Wang, Li, Jiang, Liu, and
  Schuetze}]{zhao-etal-2022-lmturk}
Mengjie Zhao, Fei Mi, Yasheng Wang, Minglei Li, Xin Jiang, Qun Liu, and Hinrich
  Schuetze. 2022.
\newblock \href {https://doi.org/10.18653/v1/2022.findings-naacl.51} {{LMT}urk:
  Few-shot learners as crowdsourcing workers in a language-model-as-a-service
  framework}.
\newblock In \emph{Findings of the Association for Computational Linguistics:
  NAACL 2022}, pages 675--692, Seattle, United States. Association for
  Computational Linguistics.

\bibitem[{Ziems et~al.(2024)Ziems, Held, Shaikh, Chen, Zhang, and
  Yang}]{ziems2023large}
Caleb Ziems, William Held, Omar Shaikh, Jiaao Chen, Zhehao Zhang, and Diyi
  Yang. 2024.
\newblock \href {https://doi.org/10.1162/coli_a_00502} {{Can Large Language
  Models Transform Computational Social Science?}}
\newblock \emph{Computational Linguistics}, pages 1--55.

\end{thebibliography}
\bibliographystyle{acl_natbib}

\newpage
\appendix

\section{Appendix}

\begin{figure*}[t!]
\begin{framed}
We will show you the caption for a photo. We will not show you the photo. Using only the caption and what you know about the world:
\begin{itemize}
\item Write one alternate caption that is \textbf{definitely} a \textbf{true} description of the photo. \textit{Example: For the caption ``\textit{Two dogs are running through a field.}'' you could write ``\textit{There are animals outdoors.}"}
\item Write one alternate caption that \textbf{might be} a \textbf{true} description of the photo. \textit{Example: For the caption ``\textit{Two dogs are running through a field.}" you could write ``\textit{Some puppies are running to catch a stick.}"}
\item Write one alternate caption that is \textbf{definitely} a \textbf{false} description of the photo. \textit{Example: For the caption ``\textit{Two dogs are running through a field.}" you could write ``\textit{The pets are sitting on a couch.}" This is different from the} maybe correct \textit{category because it's impossible for the dogs to be both running and sitting.}
\end{itemize}

In response to the original caption, please return the 3 alternate captions in a JSON readable format and include no other commentary.\\\\ \textit{Here is an example of the correct format of response to the prompt:} 

Original caption: "Two dogs are running through a field" 
\\ Three JSON-parseable alternate captions, with "definitely true", "might be true", and "definitely false" descriptions of the photo:\\ \{"true": "There are animals outdoors.",\\"maybe": "Some puppies are running to catch a stick.",\\"false": "The pets are sitting on a couch." \}
\\\\
Now, please generate the 3 alternate captions following the JSON-parseable format described earlier: Original Caption: \textbf{[INSERT SNLI PREMISE]} \\Three JSON-parseable alternate captions, with "definitely true", "might be true", and "definitely false" descriptions of the photo:
\end{framed}

\caption{\label{instructions-1}The prompt provided to all LLMs. The first four paragraphs are identical to those provided to MTurk workers for the SNLI dataset.}
\end{figure*}

\label{sec:appendix}

\begin{table*}[bth!]
\begin{adjustbox}{width=\textwidth}

\centering
\subfloat[][entailment]{
\begin{tabular}[h]{l|c c c}

\toprule
& \textbf{Word} & \textbf{$p(l|w)$} & \textbf{Freq} \\
\midrule

& Humans & 0.95 & 128 \\
& least & 0.92 & 78 \\
& activity & 0.83 & 47 \\
& multiple & 0.81 & 37 \\
& interacting & 0.85 & 34 \\
SNLI & motion & 0.97 & 32 \\
& physical & 0.83 & 30 \\
& occupied & 0.8 & 15 \\
& balances & 0.82 & 11 \\
& consuming & 0.8 & 10 \\
\midrule 
& person & 0.81 & 22264 \\
& People & 0.86 & 7059 \\
& standing & 0.84 & 4359 \\
& \text{outdoors}\textsuperscript{*} & 0.93 & 2390 \\
& engaging & 0.94 & 1689 \\
Llama & Three & 0.92 & 1593 \\
& gathered & 0.93 & 1513 \\
& activity & 0.83 & 1412 \\
& public & 0.82 & 1230 \\
& vehicle & 0.87 & 1185 \\
\midrule

& \text{There}\textsuperscript{*} & 0.99 & 16707 \\
& \text{outdoors}\textsuperscript{*} & 0.87 & 1055 \\
& three & 0.83 & 720 \\
& four & 0.88 & 335 \\
& urban & 0.83 & 318 \\
Mistral & consuming & 0.94 & 217 \\
& multiple & 0.83 & 211 \\
& vertical & 0.84 & 182 \\
& acrobatic & 0.88 & 176 \\
& many & 0.87 & 153 \\
\midrule
& person & 0.85 & 11764 \\
& \text{outdoors}\textsuperscript{*} & 0.97 & 8182 \\
& individual & 0.96 & 4569 \\
& individuals & 0.89 & 3878 \\
& \text{There}\textsuperscript{*} & 0.86 & 3794 \\
GPT-4 & Individuals & 0.97 & 2159 \\
& interacting & 0.98 & 1377 \\
& activity & 0.97 & 1250 \\
& gathered & 0.88 & 1248 \\
& public & 0.85 & 976 \\
\bottomrule
\end{tabular}
} 
\subfloat[][neutral]{
\begin{tabular}[h]{c c c}

\toprule
\textbf{Word} & \textbf{$p(l|w)$} & \textbf{Freq} \\
\midrule

tall & 0.85 & 418 \\
sad & 0.81 & 322 \\
first & 0.87 & 298 \\
owner & 0.83 & 284 \\
birthday & 0.83 & 227 \\
winning & 0.88 & 186 \\
favorite & 0.88 & 180 \\
professional & 0.83 & 149 \\
vacation & 0.94 & 141 \\
win & 0.86 & 140 \\
\midrule
Someone & 1 & 4092 \\
trying & 0.9 & 3023 \\
going & 0.95 & 1604 \\
break & 0.87 & 1339 \\
fun & 0.88 & 1165 \\
practicing & 0.86 & 1142 \\
ride & 0.82 & 811 \\
or & 0.83 & 795 \\
discussing & 0.88 & 720 \\
\text{catch}\textsuperscript{*} & 0.95 & 622 \\
\midrule

be & 0.97 & 5154 \\
trying & 0.8 & 4875 \\
may & 0.98 & 3815 \\
having & 0.85 & 2039 \\
going & 0.83 & 1877 \\
or & 0.86 & 1858 \\
friends & 0.95 & 1499 \\
It & 0.9 & 1486 \\
could & 0.98 & 1311 \\
fun & 0.92 & 1201 \\
\midrule

\text{to}\textsuperscript{*} & 0.85 & 7087 \\
for & 0.89 & 5791 \\
his & 0.82 & 5042 \\
friends & 0.94 & 3439 \\
enjoying & 0.85 & 2073 \\
couple & 0.81 & 1878 \\
from & 0.82 & 1823 \\
taking & 0.82 & 1093 \\
practicing & 0.87 & 1092 \\
team & 0.88 & 972 \\
\bottomrule
\end{tabular}
} 
\subfloat[][contradiction]{
\begin{tabular}[h]{c c c}

\toprule
\textbf{Word} & \textbf{$p(l|w)$} & \textbf{Freq} \\
\midrule

sleeping & 0.84 & 1747 \\
Nobody & 0.93 & 592 \\
asleep & 0.83 & 523 \\
\text{couch}\textsuperscript{*} & 0.81 & 477 \\
naked & 0.88 & 248 \\
tv & 0.81 & 207 \\
cats & 0.89 & 199 \\
TV & 0.81 & 177 \\
No & 0.93 & 134 \\
television & 0.83 & 124 \\
\midrule
celebrity & 0.92 & 2359 \\
actually & 0.94 & 2075 \\
cat & 0.9 & 1973 \\
Everyone & 0.93 & 1913 \\
adult & 0.89 & 1782 \\
fashion & 0.85 & 1766 \\
red & 0.84 & 1537 \\
signing & 0.92 & 1437 \\
autographs & 0.93 & 1398 \\
sleeping & 0.82 & 1371 \\
\midrule

\text{The}\textsuperscript{*} & 0.81 & 38491 \\
\text{sitting}\textsuperscript{*} & 0.83 & 14564 \\
bench & 0.87 & 8545 \\
not & 0.94 & 8068 \\
subject & 0.87 & 3672 \\
\text{couch}\textsuperscript{*} & 0.91 & 2330 \\
empty & 0.89 & 1433 \\
cards & 0.92 & 1171 \\
no & 0.92 & 955 \\
movie & 0.9 & 938 \\
\midrule
swimming & 0.92 & 16281 \\
pool & 0.91 & 14638 \\
reading & 0.8 & 3492 \\
book & 0.81 & 3048 \\
sleeping & 0.91 & 2326 \\
cooking & 0.84 & 2126 \\
cat & 0.9 & 1875 \\
dress & 0.8 & 1537 \\
alone & 0.94 & 1293 \\
library & 0.91 & 1274 \\
\bottomrule
\end{tabular}
} 
\end{adjustbox}
\caption{The most highly correlated words for each train set for given labels (the columns (a), (b), and (c)), thresholded to those with $p(l|w) \geq 0.8$ and ranked according to frequency. $^{*}$ indicates the word appears in the prompt.}\label{snli}
\end{table*}

\end{document}